\documentclass[letterpaper, 10 pt, conference]{ieeeconf}
\IEEEoverridecommandlockouts
\usepackage{amsmath,amssymb,amsfonts}
\usepackage{booktabs}
\usepackage{array}
\usepackage{multirow}
\usepackage{colortbl}
\usepackage{graphicx}
\usepackage{tikz}
\usetikzlibrary{arrows.meta,positioning,fit,calc}
\usepackage{xcolor}
\usepackage{url}

\newcommand{\soft}{\mathbf{F}}
\newcommand{\loss}{\mathcal{L}}

\newcommand{\gain}[2]{%
#1\,{\raisebox{-0.35ex}{%
\fontsize{5}{5}\selectfont
\itshape
\textcolor{red!80!black}{#2}%
}}%
}

\title{\LARGE \bf A Fully Differentiable Neuro-\textit{Soft}-Symbolic Framework for Perceptual Task Planning}

\author{
Hongyan Wei, Wael AbdAlmageed\\
Clemson University
}

\begin{document}
\bstctlcite{IEEEexample:BSTcontrol}
\maketitle
\begin{abstract}
Perceptual planning tasks require two key capabilities: accurately perceiving uncertain scenes and planning valid action sequences following logical rules. Conventional methods convert perception into discrete symbolic facts and then plan, discarding perceptual uncertainty and severing task-level feedback to perception. We introduce a generic, fully differentiable neuro-soft-symbolic framework that connects visual perception and task planning within a single computational graph. The framework maintains a continuous soft symbolic state, lifts domain rules into a differentiable soft-$T_P$ transition operator, and optimizes action logits over a short planning horizon. Gradients from the planning objective can also update the perception parameters, allowing task-relevant perceptual representations to be refined during planning. On Blocksworld, our method solves 40/40 LatPlan-40 tasks and 596/600 PlanBench-600 tasks, compared with 33/40 for LatPlan and 587/600 for the reasoning-model baseline, while requiring substantially less computation and time. In the perceptual-uncertainty ablation, our method improves the success rate from 59\% with frozen perception to 83\%. We further conduct task-and-motion simulations on Blocksworld scenes, providing an execution-level validation of the compatibility between
decoded task plans and downstream robotic motion execution.
\end{abstract}

\section{Introduction}

Visual task planning requires two complementary capabilities. The system must infer a structured symbolic state from visual observations, where perceptual information may be uncertain, and must generate an action sequence that satisfies the logical rules and constraints of the task. Neural networks provide effective perceptual representations but do not inherently guarantee valid symbolic reasoning~\cite{levine2016endtoend}. Conversely, symbolic planning methods provide explicit transition semantics and verifiable logical constraints~\cite{fikes1971strips,mcdermott1998pddl,helmert2006fastdownward} but typically assume that an accurate symbolic state is already available. A suitable planning framework must therefore leverage perceptual information while enforcing the task's logical rules and constraints.

\begin{figure*}[t]
\centering
\scalebox{1}[0.94]{%
    \includegraphics[
        width=\textwidth,
        trim=0 60 0 50,
        clip
    ]{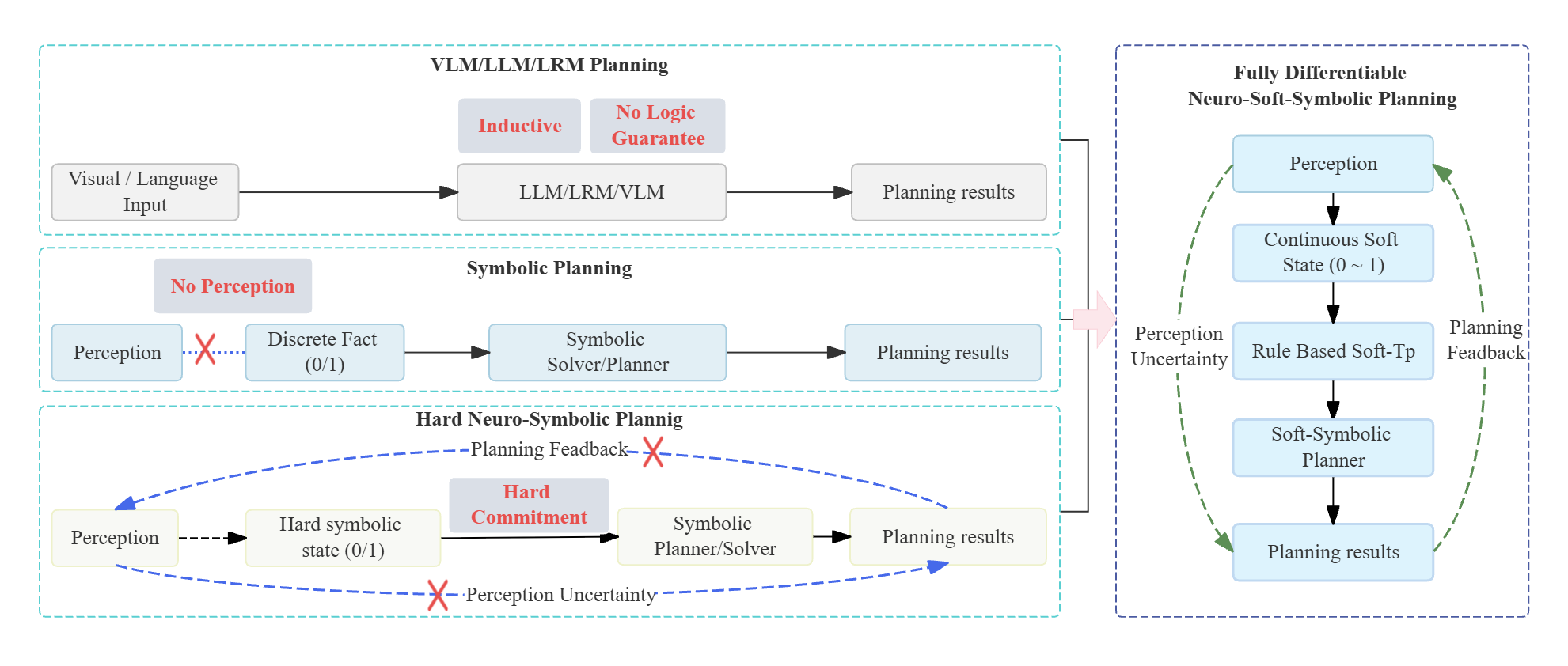}%
}
\vspace{-2mm}
\caption{Comparison of current planning methods.}
\label{fig:motivation}
\vspace{-3mm}
\end{figure*}

Existing approaches expose different limitations at the interface between perception and planning, see Figure~\ref{fig:motivation}. (1) Explicit symbolic planning operates on deterministic facts and applies a symbolic solver to generate an action sequence, but perception is outside the planning computation and the method requires an explicit, accurate symbolic state. (2) Language- and vision-language-based methods (VLM/LLM/LRM) instead infer semantic descriptions, subgoals, or candidate actions from visual or language inputs through inductive reasoning. This process can produce hallucinated or semantically inconsistent predictions and does not guarantee that the resulting actions satisfy all task constraints. (3) The hard neuro-symbolic method introduces perception into the pipeline but separates it from planning through a hard symbolic commitment: perceptual errors cannot be revised through the subsequent planning computation; the planning module cannot leverage the graded uncertainty present in the neural perceptual output; and planning objectives cannot shape the representation learned by the perception module because the symbolic handoff blocks task-level gradient feedback. 

We address this limitation with a fully differentiable neuro-soft-symbolic planning framework. Instead of committing to a single symbolic state, the framework represents perceptual relations as continuous-valued symbolic representations. These representations retain graded perceptual information while providing a structured interface for logical reasoning. Domain rules are lifted into an action-conditioned soft-$T_P$ transition operator, which propagates the soft state under candidate action distributions across a short-horizon planning window. The resulting planning objective provides gradients to the action variables and, in the perception-connected setting, to the perception parameters. In this way, the framework preserves perceptual uncertainty, enforces logical constraints during planning, and allows task-level objectives to shape the learned perceptual representation.

We evaluate the framework on visual Blocksworld planning tasks. The method solves 40/40 (100\%) LatPlan instances and 596/600 (99.33\%) PlanBench instances, compared with 33/40 (82.50\%) for LatPlan and 587/600 (97.83\%) for the PlanBench reasoning-model baseline. It requires an average of 9.96 s per PlanBench instance, compared with 40.43 s for the evaluated o1-preview LRM.

The contributions of this work are:

\begin{itemize}
\item A generic, task-agnostic, fully differentiable
neuro-soft-symbolic framework that unifies visual perception and task
planning within a single computational graph. It maintains continuous
symbolic interpretations throughout rule-guided planning, allowing
task objectives to influence both action optimization and perceptual
representations.

\item A logic-guided soft-$T_P$ transition operator that replaces hard symbolic commitment with continuous soft states, which form the basis for a fully differentiable perception--planning computation, while lifting domain rules and action schemas into differentiable temporal propagation under candidate action distributions and preserving perceptual uncertainty.

\end{itemize}

\section{Related Work}

\textbf{Classical symbolic planning.} Classical planning originated with explicit symbolic preconditions and effects~\cite{fikes1971strips} and was later standardized through languages such as PDDL~\cite{mcdermott1998pddl}. Modern planners provide efficient search, explicit transition semantics, and verifiable plans~\cite{helmert2006fastdownward}; integrated task-and-motion planning extends this formulation with geometric feasibility and motion constraints~\cite{garrett2021integrated}. These methods assume the current task state is already available as reliable symbolic facts. Consequently, perception uncertainty is either handled by a separate front end or omitted from the planning goal.

\textbf{Neuro-symbolic reasoning.} Logic Tensor Networks represent logical theories with differentiable tensor semantics~\cite{serafini2016ltn}, while DeepProbLog combines neural predicates with probabilistic logic programs~\cite{manhaeve2018deepproblog}. Neural Logic Machines further demonstrate how relational structure and logical operations can be learned with neural modules~\cite{dong2019neural_logic}. Recent soft answer-set and differentiable propagation methods provide additional foundations for retaining logical structure in continuous computation~\cite{takemura2024differentiable,eiter2026ndprop}. These works inspire our use of continuous symbolic
interpretations and differentiable rule lifts, but they primarily
address static deduction or state propagation rather than
temporal planning with execution.

LatPlan learns a propositional representation from images and performs classical planning in the learned latent space~\cite{asai2022latplan}. Model-based visual control methods learn latent dynamics and optimize imagined rollouts from pixels~\cite{ebert2018visualforesight,hafner2020dreamer}, while other approaches learn latent plans from demonstrations or play~\cite{lynch2020learning}. These methods demonstrate the value of planning over representations, but their latent states do not generally expose verifiable rule-level transition semantics. 

\textbf{Language- and vision-language-based planning.} Language models provide a semantic mechanism for task decomposition and action generation. LLM+P translates natural-language problems into PDDL and delegates search to a classical planner~\cite{liu2023llmp}, while SayCan grounds language-model suggestions in learned robotic affordances~\cite{ahn2022saycan}. Other language-conditioned systems use code generation, embodied feedback, or hierarchical task planning for robotic tasks~\cite{liang2023codepolicies,huang2023innermonologue,kwon2025fast}. PlanBench evaluates language models on planning and reasoning about change~\cite{valmeekam2023planbench}, while later evaluation shows that additional reasoning-model computation does not by itself guarantee executable and verifiable plans~\cite{valmeekam2025o1}. These methods rely on language-mediated semantic prediction, translation, parsing, or external verification, and they do not provide a differentiable route from a task objective through rule-constrained temporal state updates to a visual representation. Our method does not require an LLM/LRM, or a natural-language intermediate representation; task semantics are specified directly through domain rules.

\textbf{Task and motion planning.} Classical task-and-motion planning couples symbolic task decisions with geometric feasibility, collision avoidance, and robot motion generation~\cite{garrett2021integrated}. Differentiable task-and-motion and trajectory-optimization methods introduce gradients into parts of this process~\cite{shen2024differentiable}. Our current framework addresses the high-level task-planning component, with actions represented as relational operators such as moving an entity to a support target. This formulation provides a natural basis for future integration with differentiable motion optimizers.

Overall, existing neuro-symbolic methods primarily study static inference or rule evaluation, whereas planning methods generally operate on committed symbolic states or use non-differentiable interfaces between perception, semantics, and search. Our work addresses this gap with a fully differentiable neuro-soft-symbolic framework in which continuous symbolic representations remain connected to logical state transitions, action optimization, and the task objective within one computational graph.

\begin{figure*}[t]
\centering
\includegraphics[width=\textwidth]{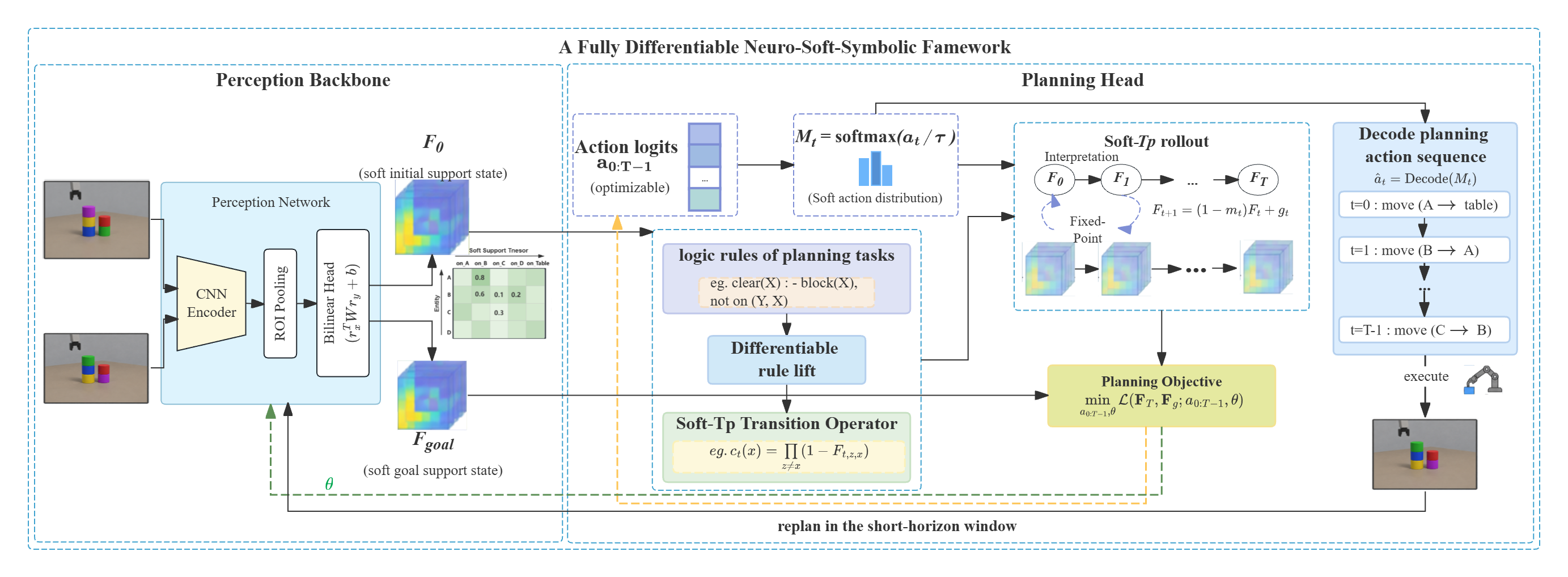}
\caption{Overview of the proposed fully differentiable neuro-soft-symbolic planning framework.}
\label{fig:framework}
\end{figure*}

\section{Method}

\subsection{Framework Overview}
Figure~\ref{fig:framework} gives the complete pipeline. The perception module maps current and goal observations to continuous support interpretations. Domain rules are then lifted into an action-conditioned soft-$T_P$ transition operator, which propagates the interpretation under soft action distributions across a short planning window. The resulting terminal state is evaluated by a planning objective, whose gradients reach the action logits and, in the perception-connected setting, the perception parameters. After optimization, the action distribution is decoded into a legal planning action sequence; execution produces the next observation and triggers the next short-horizon replanning window.

\subsection{Problem Formulation}
We consider a deterministic symbolic task represented by a domain transition relation and a visual observation pair $(I_0,I_g)$. Let $N$ entities interact with $K$ base slots. For Blocksworld, $K=1$ and the base slot is an unlimited-capacity table. A support state is
\begin{equation}
    \soft \in [0,1]^{N\times(N+K)}, \qquad \sum_y \soft_{x,y}=1,
\end{equation}
where $\soft_{x,y}$ denotes the belief that entity $x$ directly rests on target $y$.

At each planning step $t$, the optimizer maintains action logits $a_t\in\mathbb{R}^{N\times(N+K+1)}$. The extra column represents STAY. The corresponding action distribution is $M_t=\operatorname{softmax}(a_t/\tau)$. Given an initial support state and a goal support state, the short-term planning problem is
\begin{equation}
\small
\min_{a_{0:T-1},\,\theta}\loss(\soft_T,\soft_g)
\quad\text{s.t.}\quad
\soft_T=\Phi_\theta^{(T)}(\soft_0,a_{0:T-1}).
\label{eq:planning_problem}
\end{equation}
where $\theta$ denotes perception parameters in the perception-connected variant.

\subsection{Visual-to-Soft-Symbolic Perception}
The perception network uses a CNN backbone followed by region-of-interest pooling for each entity. Let $r_x$ be the embedding of entity $x$ and let $r_y$ be the embedding of a candidate support target, including a learned base-slot embedding. A directional bilinear head assigns
\begin{equation}
 s(x,y)=r_x^\top W r_y+b,
 \qquad \soft_{x,:}=\operatorname{softmax}_y(s(x,y)).
\end{equation}
The matrix $W$ is not constrained to be symmetric, because ``$x$ on $y$'' and ``$y$ on $x$'' are different relations. Self-support entries and padded entities are masked before the softmax. This produces a continuous symbolic interpretation of the image without an intermediate argmax.

\begin{table*}[t]
\centering
\caption{Correspondence between Blocksworld logic rules and the
differentiable soft-$T_P$ lift used in our implementation.}
\label{tab:rules}
\scriptsize
\setlength{\tabcolsep}{1.5pt}
\renewcommand{\arraystretch}{0.78}
\resizebox{0.92\textwidth}{!}{%
\begin{tabular}{@{}p{5.0cm}p{4.6cm}p{5.4cm}@{}}
\toprule
Logic rule / constraint & ASP syntax & Soft-$T_P$ \\
\midrule

$\operatorname{clear}(X)$: $X$ is clear if and only if no other
block currently rests on $X$.
&
\texttt{clear(X) :- block(X), not blocked(X).} \newline
\texttt{blocked(X) :- on(Y,X), Y!=X.}
&
$c_t(x)=\prod_{z\ne x}(1-\soft_{t,z,x})$
(Eq.~\ref{eq:clear}) --- product-t-norm lift. \\

\midrule

$\mathrm{move}(x\!\to\!y)$ is allowed only when $x$ and $y$ are
available.
&
\texttt{legal(X,Y) :- clear(X), clear(Y), X!=Y.}
&
$\tilde g_t(x,y)
=M_t(x,y)c_t(x)\bar c_t(y)e_t(x,y)$
(Eq.~\ref{eq:raw_gate}). \\

\midrule

At most one block moves at a given timestep.
&
\texttt{:- moves(X,T), moves(Z,T), X!=Z.}
&
$\rho_t(x)=\sum_y\tilde g_t(x,y)$,
$q_t(x)=\prod_{x'\ne x}(1-\rho_t(x'))$, and
$g_t(x,y)=\tilde g_t(x,y)q_t(x)$
(Eqs.~\ref{eq:competition}--\ref{eq:committed_gate}). \\

\midrule

A block supports at most one other block; the table has unlimited
capacity.
&
\texttt{:- on(X,Y), on(Z,Y), X!=Z, block(Y).}
&
$\mathcal{T}_{P}^{\mathrm{targ}}(\soft)_{z,y}
=\soft_{z,y}\prod_{z'\ne z}(1-\soft_{z',y})$
(Eq.~\ref{eq:target_exclusivity}). \\

\midrule

A block remains on its current support unless a move commits.
STAY creates no new relation.
&
\texttt{on(X,Y,T+1) :- on(X,Y,T),} \newline
\texttt{not moved(X,T).}
&
$\soft_{t+1,x,y}
=(1-m_t(x))\soft_{t,x,y}+g_t(x,y)$,
where $m_t(x)=\sum_y g_t(x,y)$
(Eq.~\ref{eq:transition}). \\

\bottomrule
\end{tabular}%
}
\end{table*}

\subsection{Differentiable Soft-$T_P$ Transition}

The soft-$T_P$ transition operator provides the domain-rule
interface for the shared differentiable planning machinery. We
describe its Blocksworld instantiation. Each block has one support,
a block target supports at most one block, the table has unlimited
capacity, and only a clear block may move. STAY is represented
explicitly in the action distribution.

Let $\mathcal{Y}$ denote the physical support targets, excluding
STAY. The soft clear value of block $x$ is
\begin{equation}
c_t(x)
=
\prod_{z\neq x}
\left(1-\soft_{t,z,x}\right).
\label{eq:clear}
\end{equation}

For a candidate move $x\!\rightarrow\!y$, we first compute the
raw legality-gated move mass
\begin{equation}
\tilde g_t(x,y)
=
M_t(x,y)\,
c_t(x)\,
\bar c_t(y)\,
e_t(x,y),
\label{eq:raw_gate}
\end{equation}
where $\bar c_t(y)=c_t(y)$ for block targets and
$\bar c_t(y)=1$ for the table. The term $e_t(x,y)$ masks
self-support and other domain-specific invalid targets. Because at most one block move at each step, we compute
\begin{equation}
\rho_t(x)
=
\sum_{y\in\mathcal{Y}}\tilde g_t(x,y),
\qquad
q_t(x)
=
\prod_{x'\neq x}\left(1-\rho_t(x')\right),
\label{eq:competition}
\end{equation}
and define the committed move mass as
\begin{equation}
g_t(x,y)
=
\tilde g_t(x,y)\,q_t(x),
\qquad
m_t(x)
=
\sum_{y\in\mathcal{Y}}g_t(x,y).
\label{eq:committed_gate}
\end{equation}
This provides differentiable competition among move proposals.

The support state is then updated by
\begin{equation}
\soft_{t+1,x,y}
=
\left(1-m_t(x)\right)\soft_{t,x,y}
+
g_t(x,y),
\qquad y\in\mathcal{Y}.
\label{eq:transition}
\end{equation}
The explicit STAY probability and all move mass rejected by the
logical gates remain in the residual mass $1-m_t(x)$, thereby
preserving the current support distribution. Thus,
$\soft_{t+1}=\Phi(\soft_t,M_t)$ is differentiable with respect to
both the soft state and the action logits.

For block targets, we optionally enforce target-capacity
consistency with
\begin{equation}
\mathcal{T}_{P}^{\mathrm{targ}}(\soft)_{z,y}
=
\soft_{z,y}
\prod_{z'\neq z}\left(1-\soft_{z',y}\right),
\label{eq:target_exclusivity}
\end{equation}
followed by row normalization and fixed-point iteration.

The equations above instantiate the proposed framework
for Blocksworld and provide a concrete example of how domain rules are
implemented. The overall framework is generic: across
domains, the action-logit optimization, soft temporal rollout, planning
objective, gradient propagation, decoding procedure, and short-horizon
replanning remain unchanged. Only the rule-specific transition
predicates and action schemas used to instantiate $\Phi$ are replaced.
Thus, the same framework can be applied to different planning domains
without redesigning the optimization procedure.

\subsection{Differentiable Perception--Planning Optimization}
For each short-horizon window, the action logits and the perception parameters $\theta$ are updated through the same differentiable planning objective. The visual states are recomputed from the raw images at every gradient step, ensuring that each parameter update is used by the next rollout.

In addition to the goal, movement, and cycle terms, we penalize rolled-out states that have not yet reached a fixed point of the target-exclusivity operator $T_P$.
\begin{align}
 \loss_{fp} = \sum_{t=1}^{T} \bigl\| \soft_t - T_P(\soft_t) \bigr\|^2.
 \label{eq:loss_fp}
\end{align}
$\loss_{fp}$ is evaluated on each rolled-out state \emph{before} the fixed-point refinement is applied, so it directly rewards a transition that is already close to self-consistent, rather than relying on refinement alone. The full planning objective, shared by every variant, is
{\small
\begin{align}
 \loss ={}& \loss_{goal}(\soft_T,\soft_g)
 +\lambda_m\loss_{move}+\lambda_c\loss_{cycle}+\lambda_{fp}\loss_{fp}.
\label{eq:loss}
\end{align}
}
The goal term is a cross-entropy for partial-goal tasks. The movement term favors parsimonious plans, the cycle term penalizes direct two-cycles, and the fixed-point term encourages a self-consistent rollout.
For the perception-connected variant, task-level gradients also update
the perception parameters $\theta$. To limit deviation from the initial
perceptual predictions, we use KL anchors:
\begin{equation}
\loss_{PC}
=
\loss+\lambda_{anc,0}D_0+\lambda_{anc,g}D_g,
\label{eq:loss_pc}
\end{equation}
where
$D_0=D_{KL}(\soft_0\|\soft_0^{(0)})$ and
$D_g=D_{KL}(\soft_g\|\soft_g^{(0)})$.
$\soft_0^{(0)}$ and $\soft_g^{(0)}$ are the reference soft states,
while $\soft_0=P_\theta(I_0)$ and $\soft_g=P_\theta(I_g)$ remain
functions of $\theta$. Thus, the anchors regularize perception updates
without breaking the differentiable perception--planning pathway.

\begin{table*}[t]
\centering
\caption{Main planning results. All reported Ours results on Blocksworld use 3 independent planning seeds. \textcolor{red!80!black}{Red} shows the improvement of our method over the corresponding baseline.}
\label{tab:main}
\footnotesize
\setlength{\tabcolsep}{3pt}
\begin{tabular}{@{}llcccc@{}}
\toprule
Dataset & Method & Valid (SR) $\uparrow$ & Avg. time $\downarrow$ & Cost / 100 (USD) $\downarrow$ & Optimal $\uparrow$ \\
\midrule
\multirow{2}{*}{\cellcolor{white}Blocksworld, LatPlan-40}
& LatPlan ($\mathrm{AMA}^{+}_{4}$)~\cite{asai2022latplan}
& 33/40 (82.50\%) & -- & -- & -- \\

& \cellcolor{gray!12}Ours
& \cellcolor{gray!12}\gain{40/40 (100.00\%)}{+17.50\%}
& \cellcolor{gray!12}3.53 s $\pm$ 0.01\,s
& \cellcolor{gray!12}\$0.31
& \cellcolor{gray!12}31/40 (77.50\%) \\

\cmidrule(lr){1-6}

\multirow{5}{*}{\cellcolor{white}Blocksworld, PlanBench-600}
& Claude 3.5 Sonnet (LLM)~\cite{valmeekam2025o1}
& 329/600 (54.83\%) & -- & \$0.44 & -- \\

& LLaMA 3.1 405B (LLM)~\cite{valmeekam2025o1}
& 376/600 (62.67\%) & -- & -- & -- \\

& o1-mini (LRM)~\cite{valmeekam2025o1}
& 340/600 56.67\% & 10.84 s & \$3.69 & -- \\

& o1-preview (LRM)~\cite{valmeekam2025o1}
& 587/600 (97.83\%) & 40.43 s & \$42.12 & -- \\

& \cellcolor{gray!12}Ours
& \cellcolor{gray!12}\gain{596/600 (99.33\%)}{+1.50\%}
& \cellcolor{gray!12}\gain{9.96 $\pm$ 0.31\,s}{-30.47\,s}
& \cellcolor{gray!12}\gain{\$0.89}{-\$41.23}
& \cellcolor{gray!12}58.78\% $\pm$ 0.08\% \\

\midrule
\bottomrule
\end{tabular}

\vspace{1mm}

\parbox{\textwidth}{\scriptsize Our Cost is estimated from measured GPU time (9.96 s/instance, measured while running our 3 evaluation seeds concurrently on a single A100) assuming on-demand cloud A100 pricing (\$2.745/GPU-hour, AWS EC2 p4d.24xlarge, and \$3.673/GPU-hour, Google Cloud a2-highgpu-1g; both accessed September 2026).}
\end{table*}

The resulting gradients update the action logits, and, for the perception-connected variant, the perception parameters, inside the same short-term window. The action-logit learning rate is $0.1$ and the perception learning rate is $10^{-4}$. Independently of the $\loss_{fp}$ term above, we also iteratively refine each rolled-out state toward the same fixed point before the next transition step, which improves optimization stability for longer windows.

\subsection{Short-Horizon Planning and Replanning}

The optimizer plans over a short window, decodes the planning action sequence, and executes it against the current real symbolic state. The next window starts from the resulting state, following an iterative short-horizon planning procedure also used by learned latent-world-model planners~\cite{maes2026leworldmodel}. Within each short-term planning window, the forward computation, planning objective, and gradient updates form a differentiable path from the visual inputs through the soft-$T_P$ rollout to the perception parameters and action logits. The action sequence is decoded, legally executed under the task rules, and used to initialize the next planning window.

\begin{table*}[t]
\centering
\caption{PlanBench Blocksworld error analysis. The PlanBench baseline results are reported overall; our method is additionally broken down by number of blocks. Optimal is reported as a percentage of valid plans.}
\label{tab:blocks}
\footnotesize
\setlength{\tabcolsep}{3pt}
\begin{tabular}{@{\extracolsep{3pt}}ccccccc@{}}
\toprule
Blocks & Method & Instances & Valid plans $\uparrow$ & Optimal $\uparrow$ & Inexecutable plans $\downarrow$ & Non-goal-reaching plans $\downarrow$ \\
\midrule
All & o1-mini (LRM)~\cite{valmeekam2025o1} & 600 & 340 (56.67\%) & -- & 39.50\% & 3.83\% \\
All & o1-preview (LRM)~\cite{valmeekam2025o1} & 600 & 587 (97.83\%) & -- & 12 (2.00\%) & 1 (0.17\%) \\
\midrule
\rowcolor{gray!12}
Total & Ours & 600 & 596 (99.33\%) & 350 (58.78\% $\pm$ 0.08\%) & 0 (0.00\%) & 4 (0.67\%) \\
\cmidrule(lr){1-7}
3-block & Ours & 100 & 100 (100.00\%) & 84 (84.00\%) & 0 (0.00\%) & 0 (0.00\%) \\
4-block & Ours & 445 & 445 (100.00\%) & 242 (54.38\%) & 0 (0.00\%) & 0 (0.00\%) \\
5-block & Ours & 55 & 51 (92.73\%) & 24 (47.06\%) & 0 (0.00\%) & 4 (7.27\%) \\
\bottomrule
\end{tabular}
\end{table*}

\begin{table*}[t]
\centering
\caption{Ablation Study for Differentiable Perception-Planning}
\label{tab:perception_ablation}
\footnotesize
\setlength{\tabcolsep}{4pt}
\begin{tabular}{@{}lcc|cc|cc@{}}
\toprule
& \multicolumn{2}{c|}{Original PlanBench-600}
& \multicolumn{2}{c|}{Clear subset (100)}
& \multicolumn{2}{c}{Blurred subset (100)} \\
Perception Variant
& Valid (SR) $\uparrow$ & Time $\downarrow$
& Valid (SR) $\uparrow$ & Time $\downarrow$
& Valid (SR) $\uparrow$ & Time $\downarrow$ \\
\midrule

Ground-truth symbolic state (PDDL)
& 596/600 & 7.02\,s
& -- & --
& -- & -- \\

Frozen perception backbone
& 596/600 & 7.63\,s
& 100/100 & 7.98\,s
& 59/100 & 10.13\,s \\

\rowcolor{gray!12}
Planning-optimized perception backbone
& 596/600 & 9.96\,s
& 100/100 & 10.22\,s
& \gain{83/100}{+24\%}
& 13.34\,s \\

\bottomrule
\end{tabular}
\end{table*}

\section{Experiments}
\subsection{Experiments Setup}
\textbf{Dataset.} We evaluate on image-based Blocksworld and on Logistics and Sokoban tasks. Blocksworld contains the 40 image instances associated with LatPlan and 600 official PlanBench instances. The 600-instance set contains 100 3-block, 445 4-block and 55 5-block problems. Experiments were run on a single NVIDIA A100.

\textbf{Baseline.} We compare with LatPlan on the 40 image-based Blocksworld instances. On the 600-instance PlanBench Blocksworld benchmark, we include the reported zero-shot results of Claude 3.5 Sonnet, LLaMA 3.1 405B, o1-mini, and o1-preview~\cite{valmeekam2025o1}. These results correspond to the best reported performance for each listed model under the evaluation setting used in PlanBench. For the Logistics and Sokoban domains, we compare with the o1-preview results reported in the same benchmark evaluation.

\textbf{Evaluation Metrics.} Plans are decoded into action sequences and validated by executing them from initial state. We report validity, optimality, inexecutable plans, non-goal-reaching plans, average solve time, and estimated computational cost. A plan is valid if every action satisfies its preconditions and the resulting final state satisfies the goal. This plan-validation evaluates the decoded actions under the task's transition rules, thereby exposing both execution failures and failures to reach the goal.

\subsection{Main Results}

Table~\ref{tab:main} reports the main results, with optimality
reported as the percentage of valid plans. All results on the
Blocksworld LatPlan-40 and PlanBench-600 dataset use three
independent planning seeds. On LatPlan-40, our neuro-soft-symbolic
perception--planning method achieves 40/40 valid solutions, including
31/40 BFS-optimal solutions, compared with 33/40 for LatPlan. On
PlanBench-600, our method achieves 596/600 valid solutions (99.33\%),
surpassing the 587/600 (97.83\%) result of o1-preview. All four
unsuccessful PlanBench instances are non-goal-reaching; none contains
an inexecutable action sequence.

Our method also provides better efficiency on the 600-instance
Blocksworld benchmark. The average solve time is reduced from
40.43\,s to 9.96\,s per instance, while the estimated cost per 100
instances decreases from \$42.12 to \$0.89. This lower latency and
computational cost are important for practical planning systems.

\subsection{Plan Validation and Distribution Analysis }
Table~\ref{tab:blocks} reports the PlanBench baseline error distribution together with our block-count breakdown. The baseline row follows the published PlanBench Blocksworld result, while the 3-block, 4-block, and 5-block rows show our method's behavior. Our four failures occur only in the most difficult 5-block tier, as expected for a longer and more coupled planning problem. More importantly, our zero inexecutable count shows that every submitted action satisfied the modeled Blocksworld preconditions under plan validation, which is important for logic-constrained and safety-critical tasks.

\begin{figure}[t]
\centering
\includegraphics[width=\columnwidth]{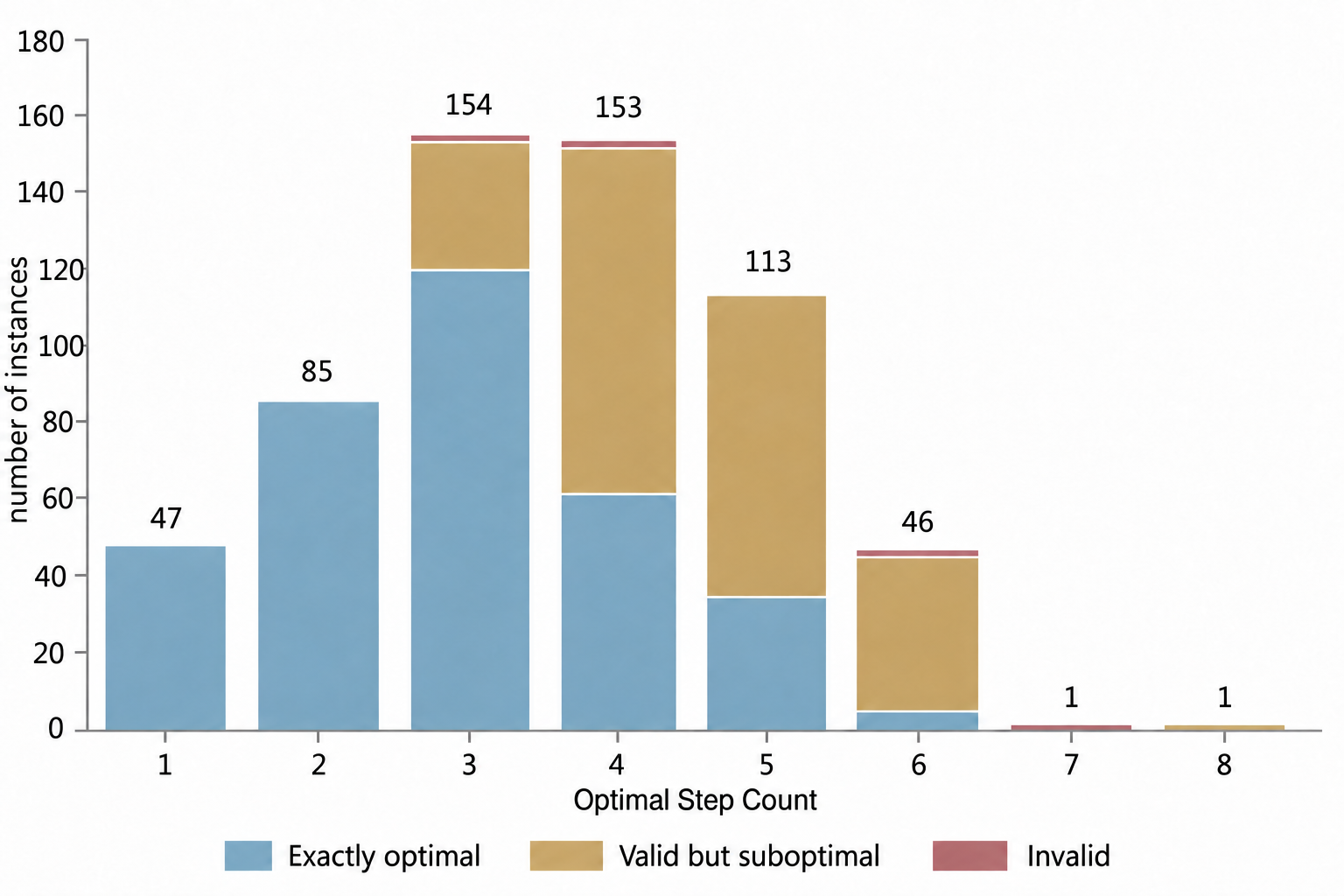}
\caption{Distribution of PlanBench-600 instances by optimal step count.}
\label{fig:stepcount}
\end{figure}

The distribution of task difficulty provides a complementary view of the PlanBench-600 results. The optimal step counts range from 1 to 8 and are concentrated between 3 and 6 steps (Figure~\ref{fig:stepcount}). Among valid-but-suboptimal instances, the mean excess is 2.47 steps, with a range of 1--7 steps. Optimal step counts on LatPlan-40 range from 0 to 4 (mean 2.12); our method matches the optimum on 31/40 instances, with a mean excess of 2.67 steps (range 1--4) among the remaining 9.

\subsection{Ablation Study for Differentiable Perception-Planning }

To evaluate our method's robustness, we conduct an ablation study on the 600-instance Blocksworld using three configurations: ground-truth symbolic state, frozen perception, and fully differentiable perception–planning. Frozen perception computes the soft symbolic state with fixed perception parameters during planning, whereas fully differentiable perception–planning allows planning gradients to update these parameters. As reported in Table~\ref{tab:perception_ablation}, on clear Blocksworld images, all three settings achieve $596/600$ valid plans. This demonstrates that our fully differentiable method—which maintains a continuous soft symbolic representation across perception and planning—reaches the performance of ground-truth symbolic inputs without requiring external solvers or reasoning models. Because Blocksworld scenes are relatively simple, the perception model nears saturation on clear images, making the benefits of updating perception parameters difficult to observe under clean inputs.
To evaluate the perception–planning connection under perceptual uncertainty, we construct a controlled image-level degradation experiment. We select a fixed 100-instance subset from PlanBench-600, preserving the original block-count proportions (17 from 3-block, 74 from 4-block, and 9 from 5-block instances). We apply a 6-pixel radius Gaussian blur to the initial and goal images before passing them to the perception backbone. All other settings (soft-$T_P$ transition, window size, perception checkpoint, seeds, and planning hyperparameters) remain unchanged, and the perception model is neither retrained nor fine-tuned on the degraded images.

On the clear 100-instance subset, both frozen perception and fully differentiable perception–planning achieve $100/100$ valid plans, confirming no difficulty bias in the subset. However, under image degradation, performance clearly diverges: frozen perception achieves $59/100$ valid plans, while our fully differentiable perception–planning method reaches $83/100$ (improved 24\%). This difference reflects how the two methods handle perceptual errors. Frozen perception treats the perceptual output from degraded images as a fixed planning input; if this output contains incorrect relational judgments, subsequent planning cannot revise them. In contrast, our method allows gradients from the planning objective to propagate through the soft symbolic state to the perception parameters, thereby correcting some perceptual errors during optimization. These results indicate that a differentiable perception–planning connection provides a positive contribution when perception is uncertain.

\subsection{Ablation Study for Short-Horizon Window Size }

The short-horizon window size used for the Blocksworld results in Table~\ref{tab:main} is
$T=3$, selected based on this ablation. We evaluate different values of $T$ on 150 independently rendered Blocksworld instances prepared for this experiment, while keeping all other hyperparameters fixed. Because the settings solve different numbers of instances, the optimal rate is computed over all 150 instances rather than only the valid plans. As shown in Table~\ref{tab:window}, $T=3$ provides the best overall trade-off, achieving the highest validity (100.0\%), the highest optimal rate (76.7\%), the lowest average time (7.77\,s), and no failures. The shorter window ($T=1$) more often terminates before reaching the goal, whereas the longer window ($T=5$) increases the cost of each optimization round and reduces validity. No setting produces an inexecutable action sequence, consistent with the zero-inexecutable result on the full 600-instance evaluation (Table~\ref{tab:blocks}).
\begin{table}[t]
\centering
\caption{Ablation study on short-horizon window-size sensitivity.}
\label{tab:window}
\small
\resizebox{\columnwidth}{!}{%
\begin{tabular}{@{}ccccccc@{}}
\toprule
$T$ & Valid (SR) $\uparrow$ & Optimal $\uparrow$ & Inexecutable $\downarrow$ & Non-goal-reaching $\downarrow$ & Avg. time $\downarrow$ \\
\midrule
1 & 108/150 (72.0\%) & 93/150 (62.0\%) & 0 (0.0\%) & 42 (28.0\%) & 9.57\,s \\
\rowcolor{gray!12}3 (ours) & 150/150 (100.0\%) & 115/150 (76.7\%) & 0 (0.0\%) & 0 (0.0\%) & 7.77\,s \\
5 & 128/150 (85.3\%) & 114/150 (76.0\%) & 0 (0.0\%) & 22 (14.7\%) & 13.48\,s \\
\bottomrule
\end{tabular}
}%
\end{table}

\begin{figure*}[t]
\centering
\includegraphics[
    width=\textwidth,
    trim=0 65 0 50,
    clip
]{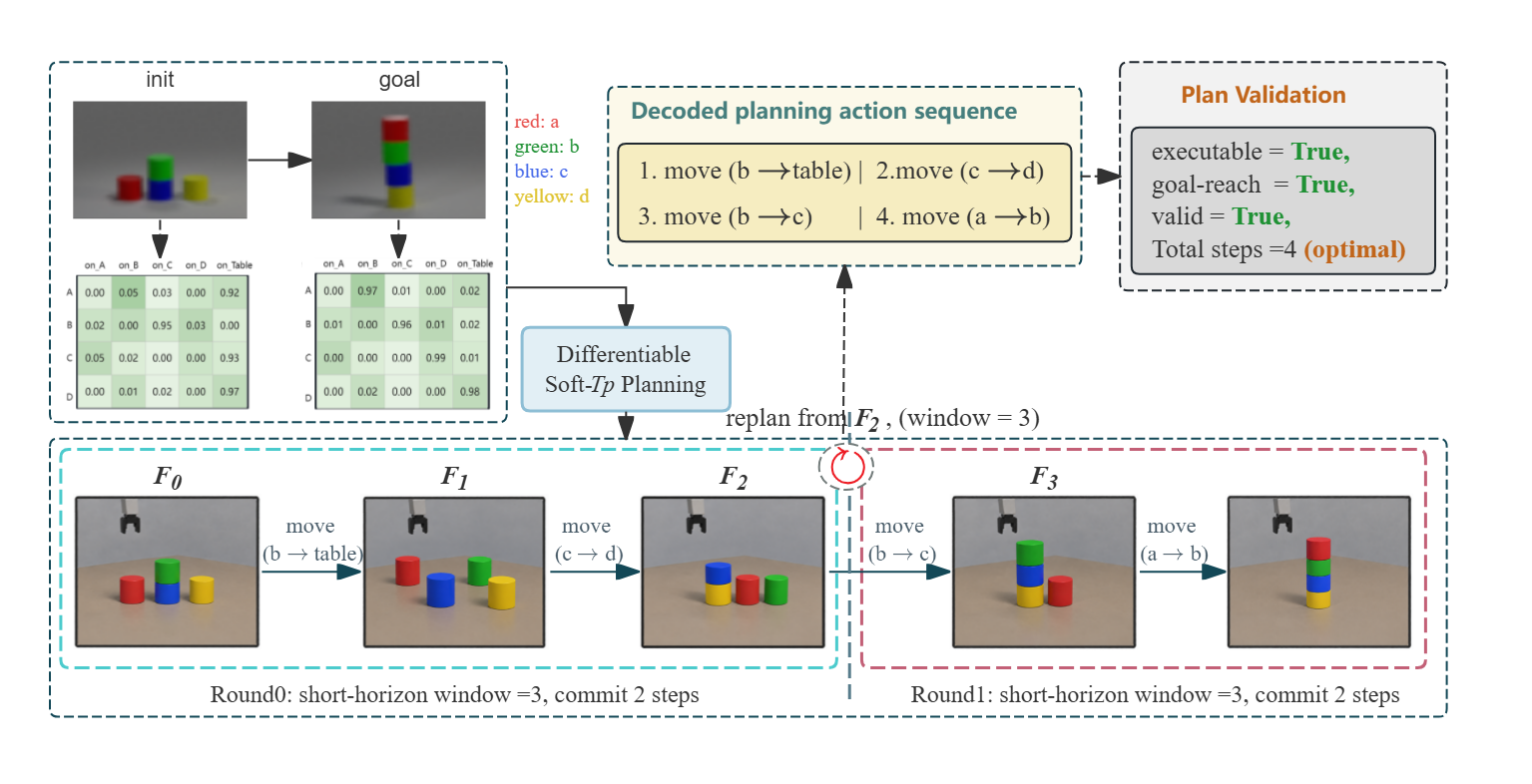}
\vspace{-2mm}
\caption{Case study of the proposed perception-to-planning procedure on a 4-block PlanBench instance.}
\label{fig:instance50}
\vspace{-3mm}
\end{figure*}

\subsection{Case Study on Blocksworld Instance}

Figure~\ref{fig:instance50} shows the complete computation for a specific four-block Blocksworld problem from PlanBench. The initial state is $a$ on the table, $b$ on $c$, $c$ on the table, and $d$ on the table. The goal state requires $a$ on $b$, $b$ on $c$, $c$ on $d$, and $d$ on the table. As shown by $F_0$, the perception module processes the initial image into soft probabilities in $[0,1]$ over continuous support relations.
 In the first round, corresponding to $t=0,1,2$, no candidate action at $t=2$ reaches the decoder confidence threshold of $0.3$ for the Blocksworld task. Therefore, only the actions at $t=0$ and $t=1$ are submitted, and the next planning round starts from the state $F_2$ obtained in the first round.

This example highlights how the differentiable rule
constraints determine action ordering. Although the relation
``$b$ on $c$'' already satisfies part of the goal, $c$ cannot
be moved to $d$ while $b$ remains on $c$. From
Eq.~\ref{eq:clear}, $F_{b,c}\approx 1$ drives $c_t(c)$ close
to zero. Consequently, the raw move mass
$\tilde g_t(c,d)$ in Eq.~\ref{eq:raw_gate} remains negligible
regardless of how strongly the action logits favor
$\mathrm{move}(c\!\rightarrow\!d)$. The optimizer must
therefore first assign mass to
$\mathrm{move}(b\!\rightarrow\!\mathrm{table})$, which
increases the subsequent clear value of $c$ and makes
$\mathrm{move}(c\!\rightarrow\!d)$ feasible.

Across the two planning windows, the planner produces the
four-action:
$\mathrm{move}(b\!\rightarrow\!\mathrm{table})$,
$\mathrm{move}(c\!\rightarrow\!d)$,
$\mathrm{move}(b\!\rightarrow\!c)$, and
$\mathrm{move}(a\!\rightarrow\!b)$. The first window commits the first two actions and reaches $F_2$. The planner then replans from $F_2$ and commits the remaining two actions in the second window. Plan validation replays the complete decoded sequence from the initial state, confirming that all four actions are executable and the final state satisfies the goal.
\begin{figure*}[t]
\centering
\includegraphics[
    width=\textwidth,
    trim=0 60 0 30,
    clip
]{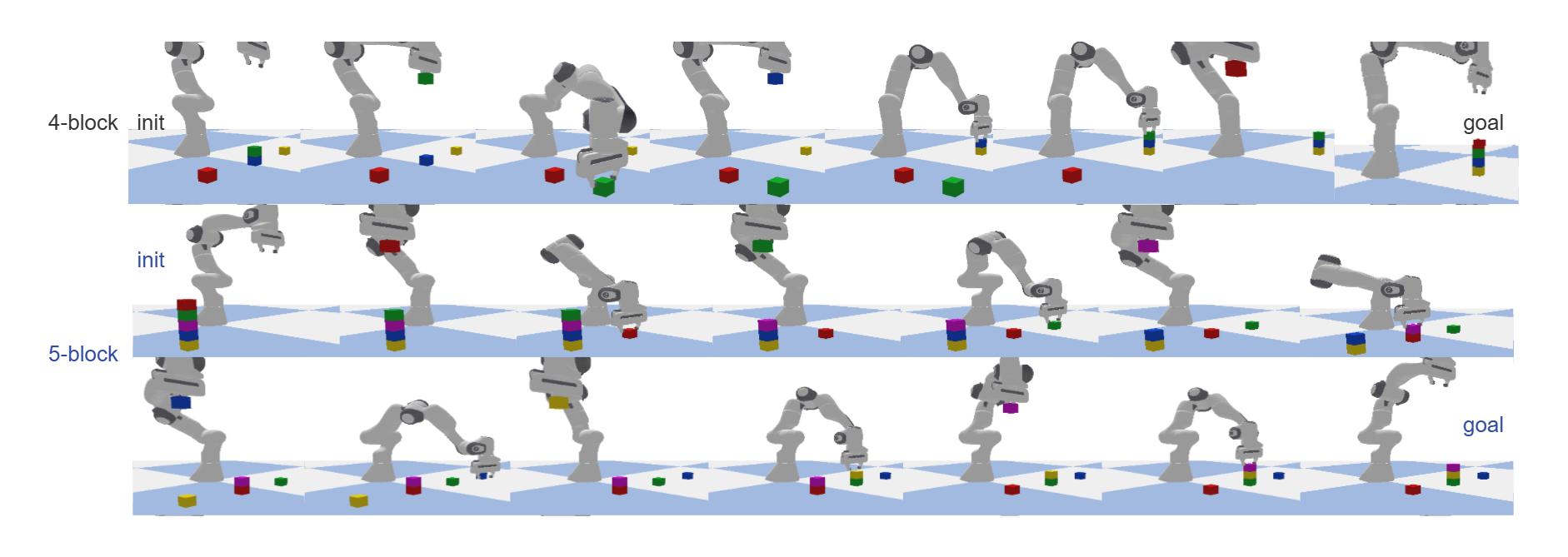}
\caption{Integrated task-and-motion simulation on four-block and
five-block Blocksworld instances.}
\label{fig:simulation}
\end{figure*}

\subsection{Task-and-Motion Simulation}
To evaluate whether the decoded task actions can be grounded in downstream motion execution, we integrated the action sequences from PlanBench Blocksworld instances into a PyBullet simulation of a Franka Emika Panda arm. Each relational action $\texttt{move}(x,y)$ is realized as a pick-and-place primitive, with inverse kinematics converting Cartesian waypoints for approach, grasp, transport, and release into the corresponding arm joint configurations. This experiment provides an execution-level validation of the task-to-motion connection between high-level task planning and downstream motion execution. The simulation results show that the relational action sequences generated by our method can be translated into concrete arm motions and successfully execute the corresponding block manipulations. Figure~\ref{fig:simulation} shows key simulation frames from two instances, including the pick-and-place operation for each action step.

\begin{table*}[t]
\centering
\caption{Cross-domain comparison on Logistics and Sokoban.}
\label{tab:domains}
\footnotesize
\setlength{\tabcolsep}{3pt}
\begin{tabular}{@{}llcccc@{}}
\toprule
Dataset & Method & Valid (SR) $\uparrow$ & Inexecutable $\downarrow$ & Non-goal-reaching $\downarrow$ & Avg. time $\downarrow$ \\
\midrule
\multirow{2}{*}{\cellcolor{white}Logistics} & o1-preview (LRM) & 188/200 (94.00\%) & 12 (6.00\%) & 0 (0.00\%) & 84.07 s \\
& \cellcolor{gray!12}Ours & \cellcolor{gray!12}\gain{196/200 (98.00\%)}{+4.00\%} & \cellcolor{gray!12}0 (0.00\%) & \cellcolor{gray!12}4 (2.00\%) & \cellcolor{gray!12}\gain{36.28 s}{-47.79\,s} \\
\cmidrule(lr){1-6}
\multirow{2}{*}{\cellcolor{white}Sokoban} & o1-preview (LRM) & 7/55 (12.73\%) & 37 (67.27\%) & 11 (20.00\%) & 147.98 s \\
& \cellcolor{gray!12}Ours & \cellcolor{gray!12}\gain{33/55 (60.00\%)}{+47.27\%} & \cellcolor{gray!12}0 (0.00\%) & \cellcolor{gray!12}22 (40.00\%) & \cellcolor{gray!12}\gain{53.02 s}{-94.96\,s} \\
\bottomrule
\end{tabular}
\end{table*}

\subsection{Cross-Domain Evaluation}
We also apply the soft-symbolic planning machinery to two additional domains of PlanBench. Logistics tests transport and fleet-routing structure, while Sokoban tests irreversible manipulation actions and deadlock-sensitive planning in constrained spaces. On the same 200-instance Logistics set, our method solves 196/200 instances (98.00\%), compared with 188/200 (94.00\%) for o1-preview. In Sokoban, our method solves 33/55 instances (60.00\%), compared with 7/55 (12.73\%) for o1-preview. In both domains, our method produces no inexecutable action sequence; the remaining failures are non-goal-reaching. The corresponding error counts and average times are reported in Table~\ref{tab:domains}. These domains use domain-specific transition rules while retaining the same differentiable action-optimization principle.

\section{Conclusion and Future Work}
We presented a generic, fully differentiable neuro-soft-symbolic
framework for perceptual task planning. It maintains continuous
symbolic interpretations, preserves perceptual uncertainty, and connects perception to planning through a differentiable soft-$T_P$ rollout within each optimization window. On the Blocksworld benchmark, our method achieves higher planning validity than the baselines, while substantially reducing solve time and estimated cost relative to the LLM/LRM baseline. These efficiency gains provide a practical advantage for embodied planning systems operating under latency and resource constraints. The perceptual-uncertainty ablation further shows that the differentiable perception--planning connection improves robustness to perceptual uncertainty.

The current evaluation focuses on compact relational domains with a manageable number of entities and action types. In future work, we will extend the framework to more demanding settings with many interacting entities, long chains of dependencies, richer action schemas, partial observability, and geometric constraints. These extensions require scalable relational representations, structured decomposition, and efficient rule propagation to broaden the framework's applicability to embodied AI and robotic planning while preserving the differentiable perception--planning connection.

\bibliographystyle{IEEEtran}
\bibliography{references}
\end{document}